\documentclass[letterpaper, 10 pt, conference]{ieeeconf}  
\pdfoutput=1

\IEEEoverridecommandlockouts                              

\usepackage{times} 

\usepackage[ruled,linesnumbered]{algorithm2e}
\usepackage{algpseudocode}

\usepackage{amsmath} 
\usepackage{amssymb}  
\usepackage{color}
\usepackage{cite}
\usepackage{graphicx}
\usepackage{subfigure}
\usepackage[english]{babel}
\usepackage{amsthm}
\usepackage{breqn}
\usepackage[super]{nth}
\usepackage{url}
\usepackage{siunitx}
\usepackage{makecell}
\usepackage{microtype}
\usepackage{booktabs}
\usepackage{hyperref}
\usepackage{stfloats}
\usepackage{threeparttable} 
\usepackage{multirow}
\theoremstyle{definition}

\usepackage{verbatim}

\usepackage[utf8]{inputenc}
\usepackage{graphicx}
\usepackage{float}
\usepackage{bm}
\usepackage{amssymb}
\usepackage{threeparttable}
\usepackage{multirow}
\usepackage{eqnarray}
\usepackage{color}
\usepackage{graphicx}
\usepackage{tablefootnote}

\makeatletter
\newcommand{\removelatexerror}{\let\@latex@error\@gobble}
\makeatother

\SetKwInOut{KwParameter}{Parameters}

\title{\LARGE \bf
Contact-Implicit Model Predictive Control for Dexterous\\
In-hand Manipulation: A Long-Horizon and Robust Approach
}

\author{Yongpeng Jiang,
Mingrui Yu,
Xinghao Zhu,
Masayoshi Tomizuka,
and Xiang Li
\thanks{
Y. Jiang, M. Yu, and X. Li are with the Department of Automation, Beijing National Research Center for Information Science and Technology, Tsinghua University, China. X. Zhu and M. Tomizuka are with the Department of Mechanical Engineering, University of California, Berkeley,
CA, USA. This work was supported in part by the National Key R\&D Program of China under Grant 2022YFB4701403, in part by the National Natural Science Foundation of China under Grant U21A20517, 52075290, and 623B2059, and in part by the Institute for Guo Qiang, Tsinghua University.
Corresponding author: Xiang Li (xiangli@tsinghua.edu.cn).
}}

\begin{document}

\maketitle
\thispagestyle{empty} 
\pagestyle{empty}  

\newtheorem{definition}{Definition}

\begin{abstract}

Dexterous in-hand manipulation is an essential skill of production and life. However, the highly stiff and mutable nature of contacts limits real-time contact detection and inference, degrading the performance of model-based methods. Inspired by recent advances in contact-rich locomotion and manipulation, this paper proposes a novel model-based approach to control dexterous in-hand manipulation and overcome the current limitations. The proposed approach has an attractive feature, which allows the robot to robustly perform long-horizon in-hand manipulation without predefined contact sequences or separate planning procedures. Specifically, we design a high-level contact-implicit model predictive controller to generate real-time contact plans executed by the low-level tracking controller. Compared to other model-based methods, such a long-horizon feature enables replanning and robust execution of contact-rich motions to achieve large displacements in-hand manipulation more efficiently; Compared to existing learning-based methods, the proposed approach achieves dexterity and also generalizes to different objects without any pre-training. Detailed simulations and ablation studies demonstrate the efficiency and effectiveness of our method. It runs at 20Hz on the 23-degree-of-freedom, long-horizon, in-hand object rotation task.

\end{abstract}

\section{Introduction}
In-hand dexterous manipulation is an essential skill with a wide range of applications, such as turning doorknobs in the home service, operating all kinds of tools and machines on the production line, and so on. Such a skill is based on contact-rich manipulation, where the robot has to make and break contacts to interact with objects \cite{Billard2019TrendsAC}. It is challenging to control the contact-rich motions because the non-smooth contact mechanism leads to intractable dynamics and stiff loss landscapes \cite{Pang2022GlobalPF}. Although considerable effort has been devoted to the locomotion domain with great success \cite{Kim2023ContactImplicitMC}, manipulation is quite different. For example, the robot must actively explore contacts instead of relying on gravity. Also, the number and locations of the contacts are constantly changing.
Existing works adopt both learning-based and model-based approaches to address these challenges. Despite the surprising robustness of learning, probably due to domain randomization and large-scale training, its data inefficiency slows down further deployment and generalization \cite{Jin2022TaskDrivenHM, Andrychowicz2018LearningDI, Chen2022VisualDI, Bui2023EnhancingTP, Xu2022AcceleratedPL}.
In contrast, model-based techniques provide a plug-and-play solution and can be divided into two categories. The first category, the contact-explicit approach, controls the robot to make contacts at predefined locations \cite{Khadivar2023AdaptiveFC}. This transforms the manipulation problem into the discrete search for contact sequences and the continuous optimization of control inputs \cite{Chen2021TrajectoTreeTO, Cheng2023EnhancingDI, Zhu2022EfficientOM, Cruciani2018DexterousMG}. Although this allows for robust execution on hardware with simple controllers, problems can arise when dealing with long-horizon tasks. This is because manipulation with specific contacts easily falls into the local optimum, and replanning new sequences is time-consuming. The second category takes a more elegant contact-implicit approach. It is common to use smooth surrogate models \cite{Pang2022GlobalPF, Kurtz2023InverseDT} or relaxed complementary constraints \cite{Kim2023ContactImplicitMC} to efficiently explore possible contacts. 
Methods of this kind do not require specific contact sequences, and can adapt to potential contacts online instead.
However, implicit methods always sacrifice physical fidelity for computational efficiency, thus create difficulties for long-horizon tasks \cite{Kim2023ContactImplicitMC, Pang2022GlobalPF}, as the system tends to evolve differently under unexpected deviation of contacts.
\begin{figure}[t]
    \centering
    \includegraphics[width=.98\columnwidth]{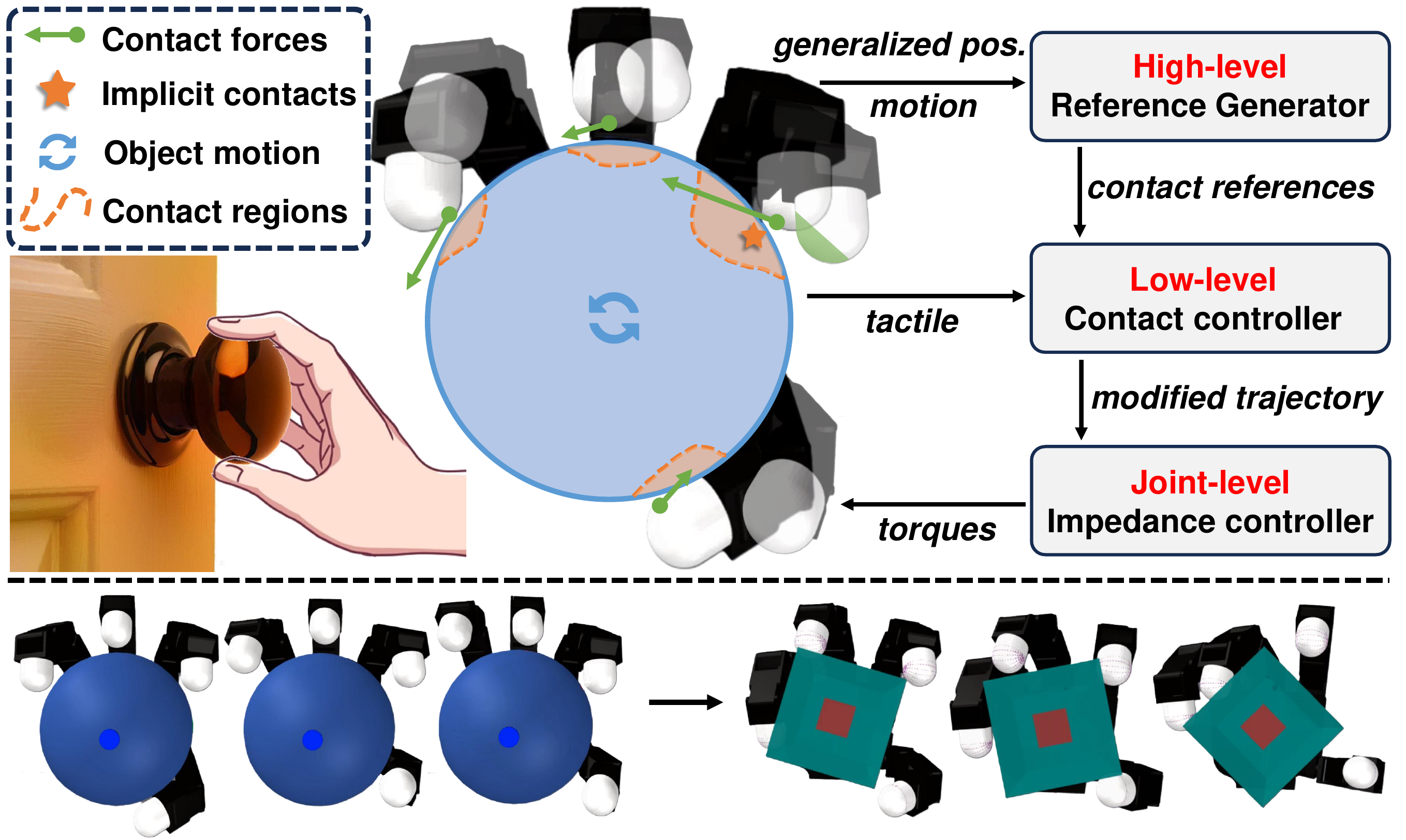}
    \vspace{-10pt}
    \caption{
        \textbf{Overview of the proposed method for long-horizon in-hand manipulation.}
        The reference generator infers how to make and break contacts in the system's neighborhood based on desired object motions. Then, the contact controller addresses modeling errors with tactile feedback and executes the inferred contacts to the fullest. Finally, joint-level impedance control is used to execute the contact-rich motions. The proposed method can seamlessly generalize among different objects.
    }
    \label{fig: teaser}
    \vspace{-15pt}
\end{figure}
In this paper, we propose a novel control formulation for long-horizon in-hand manipulation, where the fingers alternately make and break contacts to achieve a large object displacement. The contact-rich motions of the multi-finger hand are regulated by a high-level reference generator and a low-level contact controller, shown in Fig.~\ref{fig: teaser}. Specifically, the proposed method is realized with a hierarchical structure:
\begin{itemize}
    \item \textbf{At high-level}, we formulate a contact-implicit model predictive control (MPC) scheme to compute reference positions and contact forces given desired object motions. The controller runs a differential dynamic programming (DDP) algorithm on a smooth, differentiable quasi-dynamic model.
    \item \textbf{At low-level}, we formulate a simple controller to track the high-level references. The controller incorporates a compliant contact model and utilizes tactile feedback.
\end{itemize}
The high-level module infers potential contacts in the surrounding area that result in the desired object motions. Simultaneously, the low-level module regulates the system around high-level references to ensure the inferred contacts are executed to the fullest.
Compared to contact-explicit approaches, our method needs no pre-defined trajectories or separate planning procedures. Compared to contact-implicit approaches, our method has superior robustness since it addresses ubiquitous modeling errors. These characteristics enable our method to execute long-horizon manipulation online robustly.
In addition, our method works with only trivial warm start solutions and can seamlessly generalize to different objects without any prior training required in learning-based approaches.
Simulation results demonstrate the superior performance of the proposed method compared to existing approaches. Video and codes will be available soon on the project website \footnote{https://director-of-g.github.io/in\_hand\_manipulation/}.
%

\section{Related Works}
Dexterous in-hand manipulation refers to manipulating objects without a place-and-re-grasp process \cite{Cruciani2020BenchmarkingIM}.
In this section, we discuss existing approaches to this problem. Several illuminating works from more general contact-rich locomotion and manipulation domains are included, while we omit innovations in purely mechanical design.

\subsubsection{With Reinforcement-Learning (RL) or Imitation-Learning (IL)}
Recently, learning-based approaches have shown excellent performance and stability in dexterous in-hand manipulation \cite{Weinberg2024SurveyOL}. RL-based and IL-based methods are the mainstream.
The empirical success of RL can be attributed to its inner stochasticity \cite{Pang2022GlobalPF}, domain randomization, flexible choice of sensing modalities \cite{Andrychowicz2018LearningDI}, and large-scale training with memory \cite{Chen2022VisualDI}.
In contrast, IL-based methods require less exploration of the action space but rely heavily on expert demonstrations \cite{Qin2022FromOH}.
Some methods use model-driven learning \cite{Kumar2016OptimalCW, Weinberg2024SurveyOL} or explore the combination with non-learning approaches \cite{Jin2022TaskDrivenHM, Bui2023EnhancingTP, Xu2022AcceleratedPL}.
Nevertheless, learning-based methods rely mainly on data to understand the mechanism of in-hand manipulation, making it difficult to capture the underlying commonalities from a theoretical perspective. Therefore, such methods may perform poorly for unseen objects or in the presence of significant sim2real gaps.

\subsubsection{Model-Based Design Using Explicitly Represented Contacts}
Model-based techniques provide a solution for deployment and generalization without training.
In dexterous in-hand manipulation, planning and control commonly involve explicit representation of contacts, including their locations, modes, reaction forces, and so on \cite{Cheng2023EnhancingDI}. These items form the contact sequences.
Existing works use different methods to obtain such discrete sequences, such as searching \cite{Chen2021TrajectoTreeTO, Zhu2022EfficientOM, Cruciani2018DexterousMG}, sampling \cite{Cheng2023EnhancingDI, Dafle2018InHandMV}, or demonstration \cite{Khadivar2023AdaptiveFC}.
The subsequent procedure is simplified to solving for the continuous control inputs, such as joint torques, while following pre-defined contact sequences \cite{Chen2021TrajectoTreeTO}. Thus, low-level control typically involves solving tractable optimization problems with force closure and non-sliding constraints.
While these methods are easier and more precise to implement on hardware, acquiring contact sequences adds complexity and degrades real-time performance. Moreover, if contact states deviate from pre-defined ones due to disturbance, the specific contact sequence can become misleading information.

\subsubsection{Model-Based Design Using Implicitly Encoded Contacts}
Several works avoid the increasing complexity of explicit contact representation using relaxed complementary constraints \cite{Kim2023ContactImplicitMC, Cleach2021FastCM}, smooth surrogate models \cite{Pang2022GlobalPF, nol2018ContactImplicitTO}, or direct sampling of the control \cite{Howell2022PredictiveSR}.
Aydinoglu et al. \cite{Aydinoglu2023ConsensusCC} proposed an operator splitting framework to decouple variables across time steps and accelerate the intractable Linear Complementary Problem (LCP).
Smoothing is discovered as an important procedure to allow efficient exploration among contact modes and to avoid falling into the local minimum \cite{Kim2023ContactImplicitMC, Cleach2021FastCM}. Pang et al. \cite{Pang2022GlobalPF} proved that randomized and analytic smoothing are equally important for dexterous manipulation planning.
Another method of obtaining smooth dynamics is through compliant contact models \cite{nol2018ContactImplicitTO}. Inspired by the Hunt and Crossley model, Kurtz et al. \cite{Kurtz2023InverseDT} proposed an alternative version with regularized friction and propagated the gradients through inverse dynamics. Although their method works well in the dexterous manipulation domain, the performance can be easily affected by model parameters.
Although contact-implicit methods improve efficiency by eliminating the dependence on specific contact sequences, the smoothing procedure introduces the side effect of force-at-a-distance \cite{Pang2022GlobalPF}. As a result, these methods often fail for even small discrepancies between actual and planned contact modes.
In contrast, our method is more robust due to contact feedback.
\vspace{-0.01cm}

\section{Method}
\begin{figure}[!t]
    \centering
    \includegraphics[width=1.0\linewidth]{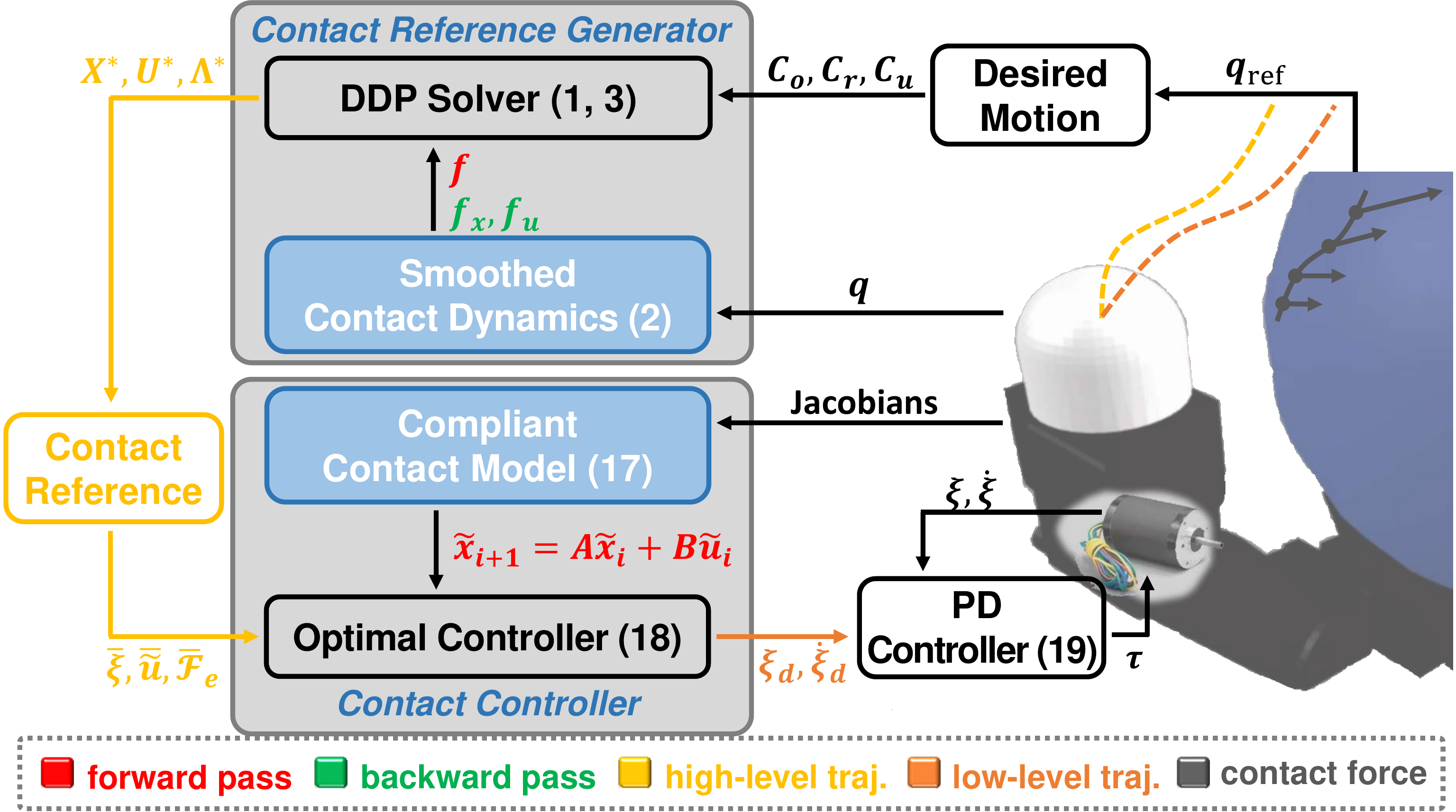}
    \vspace{-15pt}
    \caption{\textbf{Block diagram of the proposed method.} The long-horizon contact-rich manipulation is accomplished by the contact reference generator and the contact controller working as a hierarchical structure.}
    \label{fig: block_diagram_proposed_method}
    \vspace{-20pt}
\end{figure}

This paper considers the quasi-dynamic manipulation through rigid frictional contacts \cite{Mason2001MechanicsOR}. The object being manipulated is modeled as a single rigid body, while the manipulator can be any multi-link system, e.g., a multi-finger hand.
The task is defined as moving the object
to follow reference motions. The method is illustrated in Fig.~\ref{fig: block_diagram_proposed_method}.
‘Quasi-dynamic’ refers to manipulation under negligible inertia effects, in which the momentum does not accumulate and the system can be described with first-order dynamics \cite{Pang2022GlobalPF, Jiang2023ContactAwareNM}. We make such an assumption because it accords with many objects of daily life due to frictional surfaces, damped hinges, or slow motions. In this paper, we use subscripts $o$, $r$ for object and robot, and $\Vert \cdot \Vert_{\bm{W}}$ for weighted norm unless otherwise stated. Besides, $\left[\bm{a}; \bm{b}\right]$ denotes the vertical concatenation of vectors $\bm{a}, \bm{b}$, and $\left[\bm{A};\bm{B}\right]$ denotes the vertical stacking of matrices $\bm{A}, \bm{B}$.

\subsection{Contact-Implicit Reference Generator}
\label{sec: reference_generator}

%
One key issue to achieve long-horizon in-hand manipulation is to auto-generate contact references online, i.e., hand configurations, contact locations, reaction forces, and so on for the low-level controller. The process accepts only desired object motion as input without pre-defined sequences or trajectories.
\subsubsection{Problem Formulation}\label{sec: high_level_problem_formulation}
The generation of references is described as a finite horizon optimal control problem.
\begin{equation}
    \begin{aligned}
        \min_{\bm{U}=\{\bm{u}_0,\cdots,\bm{u}_{N-1}\}} \quad & \sum_{i=0}^{N-1}l(\bm{x}_i,\bm{u}_i)+l_f(\bm{x}_N)\\
        \mbox{s.t.}\quad
        &\bm{x}_{i+1}=\bm{f}(\bm{x}_i,\bm{u}_i)\\
        & \underline{\bm{u}} \preceq \bm{u}_i \preceq \overline{\bm{u}},\ i=0,\dots,N-1 \\
    \end{aligned}
    \label{eq: high_level_oc_problem}
\end{equation}
For quasi-dynamic manipulation, the state variables only include generalized positions, i.e., $\bm{x}=\bm{q}$, with the dimension of $n_q$. Here $\bm{q}$ includes both the robot and the object, i.e., $\bm{q}=\left[\bm{q}_r;\bm{q}_o\right]$. Besides, control inputs are chosen as $\bm{u}=\bm{S}_r\delta\bm{q}$ with the same bounds $\underline{\bm{u}},\overline{\bm{u}}$ at different time steps, where $\bm{S}_r$ selects the actuated dimensions. We expect the dynamics $\bm{f}$ to 1) satisfy quasi-dynamic assumptions, 2) have smooth forward and backward passes, and 3) evaluate efficiently. Property 2) is indicated in \cite{Kim2023ContactImplicitMC}, where smoothed gradients enable efficient transition among different contact modes. Moreover, we illustrate that unlike the exact integration as in \cite{Kim2023ContactImplicitMC}, the smooth forward pass is also indispensable for contact-implicit optimization in the manipulation domain.
\subsubsection{Differentiable Dynamics Through Contact}\label{sec: high_level_dynamics}
For the dynamics mapping $\bm{f}$, we use the convex, quasi-dynamic, differentiable contact (CQDC) model \cite{Pang2022GlobalPF}, which has the properties mentioned above. The model is exactly an optimization-based dynamics that relaxes the nonlinear contact mechanism as a Cone Complementary Problem (CCP) \cite{Lidec2023ContactMI}. Assuming there are $N_c$ point contacts with known local geometry and friction, the discrete-time dynamics can be formulated as:
\begin{equation}
    \begin{aligned}
        \min_{\bm{v}} \quad & \frac{1}{2}\bm{v}^\top\bm{Q}\bm{v}+\bm{b}^\top\bm{v} \\
        \mbox{s.t.}\quad
        & \mu_i\Vert \bm{J}_{i,t}\bm{v} \Vert_2 \leq \bm{J}_{i,n}\bm{v}+\phi_i/h,\ i=1,\dots,N_c \\
    \end{aligned},
    \label{eq: high_level_differentiable_dynamics}
\end{equation}
where $h$ is the discrete time step, $\bm{v}=\delta\bm{q}/h$, $\phi_i, \mu_i$ are the signed distance and friction coefficient of the $i^{\text{th}}$ contact, and $\bm{J}_i=\left[\bm{J}_{i,n};\bm{J}_{i,t}\right]$ is the contact Jacobian.
Besides, $\bm{Q}$ and $\bm{b}$ are known and can be computed with the object's inertia, the robot's joint stiffness and generalized forces applied on both. We refer the readers to \cite{Pang2022GlobalPF} for a more detailed derivation.
The original problem (\ref{eq: high_level_differentiable_dynamics}) is a second-order cone program \cite{Boyd_Vandenberghe_2011}, which is convex, differentiable, but non-smooth in nature. We found that incorporating non-smooth dynamics and its gradients in the optimal control problem (\ref{eq: high_level_oc_problem}) always results in poor convergence. Thus we adopt the analytically smoothed version of (\ref{eq: high_level_differentiable_dynamics}) and explain how the control performance is influenced by the degree of smoothness in Sec~\ref{sec: results}. The dual variable $\bm{\lambda}$ of (\ref{eq: high_level_differentiable_dynamics}) stands for the concatenated contact forces.
%

Finally, the smoothed dynamics $\bm{f}$ and differential terms $\bm{f}_{\bm{x}},\bm{f}_{\bm{u}}$ are obtained with the open-sourced implementation of \cite{Pang2022GlobalPF}, which relies on the differentiable collision detection from \cite{drake}. For the same reason as \cite{Pang2022GlobalPF}, only simple geometries with differentiable boundaries are used in the experiments, i.e., spheres, capsules. The CQDC model is also integrated into a numerical optimization framework in \cite{Pang2022GlobalPF} but for open-loop planning. However, close-loop control is further considered in this paper.
\subsubsection{Cost Function Design}\label{sec: high_level_cost_function}
The running and terminal cost terms are composed of three parts
\begin{equation}
    \begin{aligned}
        l(\bm{x},\bm{u}) &= C_o(\bm{x})+C_r(\bm{x})+C_u(\bm{u}) \\
        l_f(\bm{x}_N) &= C_o(\bm{x}_N)+C_r(\bm{x}_N)
    \end{aligned},
    \label{eq: high_level_differentiable_costs}
\end{equation}
the subscripts are omitted for brevity. The regulation costs
\begin{equation}
    \begin{aligned}
        C_o(\bm{x}) &= \Vert \bm{q}_o-\bm{q}_{o,\text{ref}} \Vert_{\bm{W}_o}^2 \\
        C_r(\bm{x}) &= \Vert \bm{q}_r-\bm{q}_{r,\text{ref}} \Vert_{\bm{W}_r}^2
    \end{aligned}
    \label{eq: high_level_regulation_costs}
\end{equation}
are the keys to generate contact-rich motions, where $\bm{q}_{\cdot,\text{ref}}$ are references and $\bm{W}_{\cdot}$ are weight matrices. Among them, $C_o$ guarantees desired object motions,
and $C_r$ encourages the periodic finger-gaiting by penalizing large finger displacements from nominal configurations.
This technique is widely applied in many model-based or learning-based frameworks that include rich contacts and periodic motions \cite{Kim2023ContactImplicitMC, qi2023general}.
The negligence of $C_r$ often causes the system to fall into local optimum, i.e., the hand fails to break contacts actively. 
In practice, the nominal finger configuration $\bm{q}_{o,\text{ref}}$ can be fixed as any grasping pose (i.e., decided empirically or generated by a grasping sampler) in most cases.
Besides, we find it helpful to increase $W_r$ near the end of the receding horizon.
We also find that considering control regulation terms improves numerical conditions and accelerates convergence.
\begin{equation}
    C_u(\bm{u}) = \Vert \bm{u} \Vert_{\bm{W}_u}^2
\end{equation}
Note that the framework allows other types of costs as long as they are differentiable. For example, additional costs can be introduced to encourage symmetric motions of certain fingers and to enhance contact clearance or maintenance.

\subsubsection{Generation of Contact References}\label{sec: high_level_contact_reference_generation}
With differentiable dynamics (\ref{eq: high_level_differentiable_dynamics}) and costs (\ref{eq: high_level_differentiable_costs}), the optimal control problem (\ref{eq: high_level_oc_problem}) can be efficiently solved with the control-limited version of DDP \cite{Tassa2014ControllimitedDD}.
As suggested by \cite{Kim2023ContactImplicitMC}, we omit the second order terms $\bm{f}_{\bm{xx}}, \bm{f}_{\bm{xu}}, \bm{f}_{\bm{uu}}$ since they contain derivatives with higher orders than kinematic Hessian, and have weaker effects on convergence.
Finally, the solved trajectories $\bm{X}^{*}=\{\bm{x}_0^{*},\cdots\bm{x}_N^{*}\}, \bm{U}^{*}=\{\bm{u}_0^{*},\cdots\bm{u}_{N-1}^{*}\}, \bm{\Lambda}^{*}=\{\bm{\lambda}_0^{*},\cdots,\bm{\lambda}_{N-1}^{*}\}$ are converted to desired robot configurations and contact forces, and are interpolated before being sent to the low-level controller.

\begin{figure}[!t]
    \centering
    \includegraphics[width=0.85\linewidth]{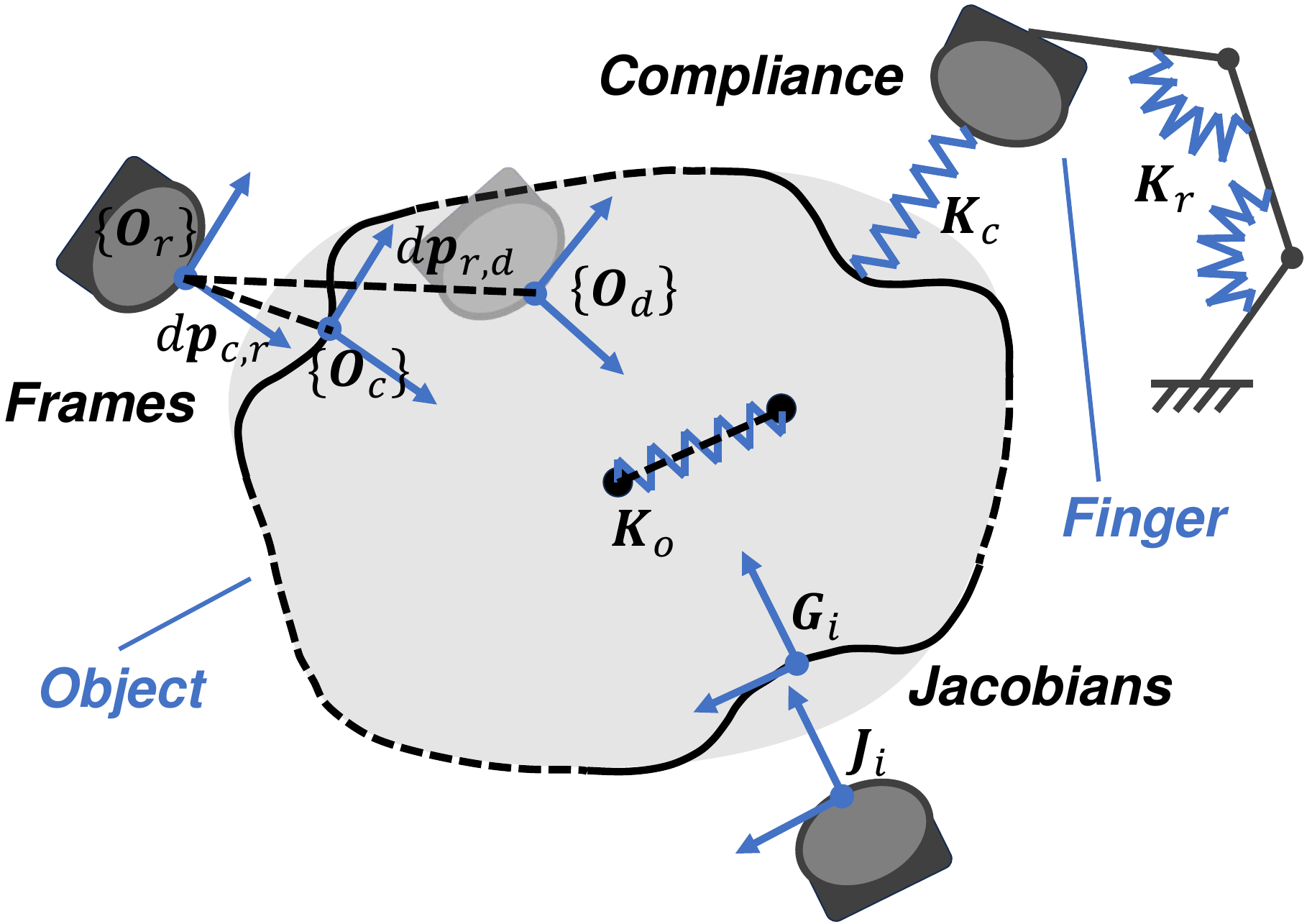}
    \vspace{-15pt}
    \caption{\textbf{Modeling of the contact controller.} The frames $\bm{O}_r,\bm{O}_c,\bm{O}_d$, frame displacements $d\bm{p}_{c,r},d\bm{p}_{r,d}$, contact jacobians $\bm{G}_i,\bm{J}_i$, stiffness  matrices $\bm{K}_c,\bm{K}_r,\bm{K}_o$ are illustrated. Note that $\bm{O}_d$ can be obtained with forward kinematics $\bm{FK}(\bm{\xi}_d)$. The model compliance creates a coupling effect between force and motion.}
    \label{fig: grasping_annotations}
    \vspace{-20pt}
\end{figure}
\subsection{Compliance-Based Contact Controller}
\label{sec: contact_controller}

%
The smoothed dynamics has side-effects that shift the object at a non-zero contact distance (i.e., force-at-a-distance). Thus directly executing the high-level references would lead to contact slippage or missing. To mitigate such modeling errors, a model-based controller is proposed to adapt the generated references locally. 
The controller incorporates an equivalent virtual spring system with a compliant contact model.
By coupling joint positions and reaction forces with compliance, a balance is achieved between tracking both. The modeling is illustrated in Fig.~\ref{fig: grasping_annotations}.
\subsubsection{Compliant Contact Model}\label{sec: low_level_compliant_model}
%
Any compliant contact model that maps relative contact distances to reaction forces can be used. One example is \cite{Kurtz2023InverseDT}, while the damping effect related to contact velocity is ignored here due to quasi-static assumptions.
The compliant contact model is guaranteed to have continuous gradients. Thus, we can define the equivalent stiffness in the contact frame,
\begin{equation}
      {^{C}\bm{K}_c}=\text{diag}\left(\frac{\partial f}{\partial \phi}[\alpha \quad \alpha \quad1]\right)
  \label{eq: contact_stiffness_matrix}
\end{equation}
where the y-axis aligns with the sliding direction, and the z-axis coincides with the contact normal, $\alpha$ is an adjustable parameter.
Note that the CQDC model is not used at low-level due to its slow, optimization-based evaluation.
\subsubsection{Contact Force-Motion Model}\label{sec: contact_force_motion_model}
We make the following assumptions: 1) quasi-dynamic manipulation, and 2) the contact stiffness and Jacobian matrices are evaluated only once at the beginning of each control step. Unless otherwise stated, we omit the subscript $i$ since the following derivation is the same for each contact. We extend the formulation in \cite{Gold2023ModelPI} to multi-contact cases.
Under quasi-dynamic assumptions, the contact force exerted by each finger can be written as
\begin{equation}
    \bm{F}_e=\bm{K}_{c} d\bm{p}_{c,r}
    \label{eq: spring_damper_external_force}
\end{equation}
where $\bm{K}_c=\bm{R}{^{C}\bm{K}_c}\bm{R}^{\top}$,
$\bm{R}$ is the rotation matrix.
Besides, $d\bm{p}_{c,r}$ is the displacement between the nearest points on the object and robot.
We attach a virtual spring at each contact with the stiffness matrix (\ref{eq: contact_stiffness_matrix}), and the equivalent stiffness $\bm{K}_o \in \mathbb{R}^{6 \times 6}$ can be obtained as
\begin{equation}
    \bm{K}_o = \sum_{i=1}^{N_c}\bm{G}_i\bm{K}_{c,i}\bm{G}_{i}^{\top}
\end{equation}
where $\bm{G}_i \in \mathbb{R}^{6 \times 3}$ equals the grasping matrix of the $i^{\text{th}}$ contact. The stacked matrix $\bm{\mathcal{G}} \in \mathbb{R}^{6 \times 3N_c}$ is defined as
\begin{equation}
    \bm{\mathcal{G}} = \left[\bm{G}_1, \cdots, \bm{G}_{N_c}\right]
\end{equation}
%

%
For clarity, $\bm{\xi}=\bm{q}_r$ is used in this section. According to (\ref{eq: joint_level_control_law}), with gravity compensation, 
it is obtained that
\begin{equation}
    \bm{K}_r(\bm{\xi}_{d}-\bm{\xi})+\bm{D}_r(\dot{\bm{\xi}}_{d}-\dot{\bm{\xi}})=\bm{\mathcal{J}}^\top\bm{\mathcal{F}}_e
    \label{eq: pd_controller_result_force_balance}
\end{equation}
where $\bm{\mathcal{J}}$ is the stacking of contact Jacobians and $\bm{\mathcal{F}}_e$ is the concatenated contact forces.
Typically we choose $\bm{K}_r=k_p\bm{I}, \bm{D}_r=k_d\bm{I}$ as diagonal matrices. 
Using $\bm{J}(\bm{\xi}_d-\bm{\xi}) \approx d\bm{p}_{r,d}$ and ignoring the velocity term, we can obtain
\begin{equation}
    k_{p}d\bm{p}_{r,d}=\bm{J}\bm{\mathcal{J}}^{\top}\bm{\mathcal{F}}_e
    \label{eq: joint_impedance_external_force}
\end{equation}
If we substitute (\ref{eq: spring_damper_external_force}) and (\ref{eq: joint_impedance_external_force}) into
\begin{equation}
    d\bm{p}_{c,d}=d\bm{p}_{c,r}+d\bm{p}_{r,d}
\end{equation}
and left-multiply $\bm{K}_c$, we have
\begin{equation}
    \bm{F}_e+k_p^{-1}\bm{K}_c\bm{J}\bm{\mathcal{J}}^{\top}\bm{\mathcal{F}}_e=\bm{K}_{c}d\bm{p}_{c,d}
    \label{eq: mapping_from_motion_to_contact_force}
\end{equation}
Note that
\begin{equation}
    \begin{aligned}
        d\dot{\bm{p}}_{c,d} &= \dot{\bm{p}}_d-\dot{\bm{p}}_c \\
        &= \bm{J}\dot{\bm{\xi}}_d-\bm{G}^{\top}\bm{K}_o^{-1}\bm{\mathcal{G}}\dot{\bm{\mathcal{F}}}_e
    \end{aligned}
    \label{eq: equations_between_force_and_motion_time_derivatives}
\end{equation}
Substituting (\ref{eq: mapping_from_motion_to_contact_force}) into (\ref{eq: equations_between_force_and_motion_time_derivatives}) for all contacts, we can obtain that
\begin{equation}
    \dot{\bm{\mathcal{F}}}_e 
\approx 
\left(\bm{I}+k_p^{-1}\bm{M}\bm{\mathcal{J}}^{\top}+\bm{N}\bm{K}_o^{-1}\bm{\mathcal{G}}\right)^{-1}\bm{M}\dot{\bm{\xi}}_d
    \label{eq: derivative_of_fe}
\end{equation}
where 
$\bm{M} \in \mathbb{R}^{3N_c \times n_{q_r}}$ is the stacking of $\bm{K}_c\bm{J}$ and $\bm{N} \in \mathbb{R}^{3N_c \times 6}$ is the stacking of $\bm{K}_c\bm{G}^{\top}$.
From (\ref{eq: pd_controller_result_force_balance}) it can be obtained that
\begin{equation}
    \dot{\bm{\xi}}=\dot{\bm{\xi}}_d+k_d^{-1}\left(k_p(\bm{\xi}_d-\bm{\xi})-\bm{\mathcal{J}}^{\top}\bm{\mathcal{F}}_e\right)
    \label{eq: derivative_of_q}
\end{equation}
With (\ref{eq: derivative_of_fe}) and (\ref{eq: derivative_of_q}), if we choose desired joint velocity as the control input $\tilde{\bm{u}}=\dot{\bm{\xi}}_d$, the second-order contact dynamics can be written as (and can be further discretized)
\begin{equation}
    \begin{bmatrix}
	 \dot{\bm{\xi}} \\
      \dot{\bm{\xi}}_d \\
      \dot{\bm{\mathcal{F}}}_e
    \end{bmatrix}
    \triangleq
    \begin{bmatrix}
	 \tilde{\bm{u}}+k_d^{-1}\left(k_p(\bm{\xi}_d-\bm{\xi})-\bm{\mathcal{J}}^{\top}\bm{\mathcal{F}}_e\right) \\
      \tilde{\bm{u}} \\
      \left(\bm{I}+k_p^{-1}\bm{M}\bm{\mathcal{J}}^{\top}+\bm{N}\bm{K}_o^{-1}\bm{\mathcal{G}}\right)^{-1}\bm{M}\tilde{\bm{u}}
    \end{bmatrix}
    \label{eq: quasi_dynamic_contact_dynamics}
\end{equation}
Note that (\ref{eq: quasi_dynamic_contact_dynamics}) describes the linear dynamics of $\Tilde{\bm{x}}=\left[\bm{\xi}; \bm{\xi}_d; \bm{\mathcal{F}}_e\right]$. 
In practice, one can omit terms including $\bm{K}_o$ when the object is not free-floating.
The low-level controller can be formulated as the following optimal control problem
\begin{align}
    \min_{\tilde{\bm{U}}} \  & \Vert \bm{\xi}_N-\bar{\bm{\xi}}_N \Vert_{\bm{W}_{\xi_{N}}} + \Vert \bm{\mathcal{F}}_{e,N}-\bar{\bm{\mathcal{F}}}_{e,N} \Vert_{\bm{W}_{F_N}} \nonumber \\
    & \sum_{i=0}^{N-1} \Vert \bm{\xi}_i-\bar{\bm{\xi}}_i \Vert_{\bm{W}_{\xi}} + \Vert \bm{\mathcal{F}}_{e,i}-\bar{\bm{\mathcal{F}}}_{e,i} \Vert_{\bm{W}_F} + \Vert \tilde{\bm{u}}_i-\bar{\tilde{\bm{u}}}_i \Vert_{\bm{W}_{\tilde{u}}} \nonumber \\
    \label{eq: low_level_oc_problem} \mbox{s.t.} \ 
    &\tilde{\bm{x}}_{i+1}=\bm{A}\tilde{\bm{x}}_i+\bm{B}\tilde{\bm{u}}_i,\ i=0,\dots,N-1
\end{align}
where the references 
are calculated from interpolating the generated references $\bm{X}^{*}, \bm{\Lambda}^{*}, \bm{U}^{*}$.
%


%
We use the joint-level PD controller with gravity compensation to convert the joint references $\bm{\xi}_d,\dot{\bm{\xi}}_d$ into joint torques.
\begin{equation}
    \bm{\tau}=\bm{\tau}_g+\bm{K}_r(\bm{\xi}_d-\bm{\xi})+\bm{D}_r(\dot{\bm{\xi}}_d-\dot{\bm{\xi}})+\bm{\mathcal{J}}^{\top}\bm{\mathcal{F}}_{e,ff}
    \label{eq: joint_level_control_law}
\end{equation}
where $\bm{\tau}_g$ is the generalized gravity. In addition, we add the feed-forward torque used in baseline methods and neglect inertia and Coriolis effects under quasi-dynamic assumptions.
%


\section{Results}\label{sec: results}
\begin{figure}[!t]
    \centering
    \subfigure[\textbf{RotZ}]{
        \begin{minipage}{4cm}
            \centering
            \includegraphics[width=4cm]{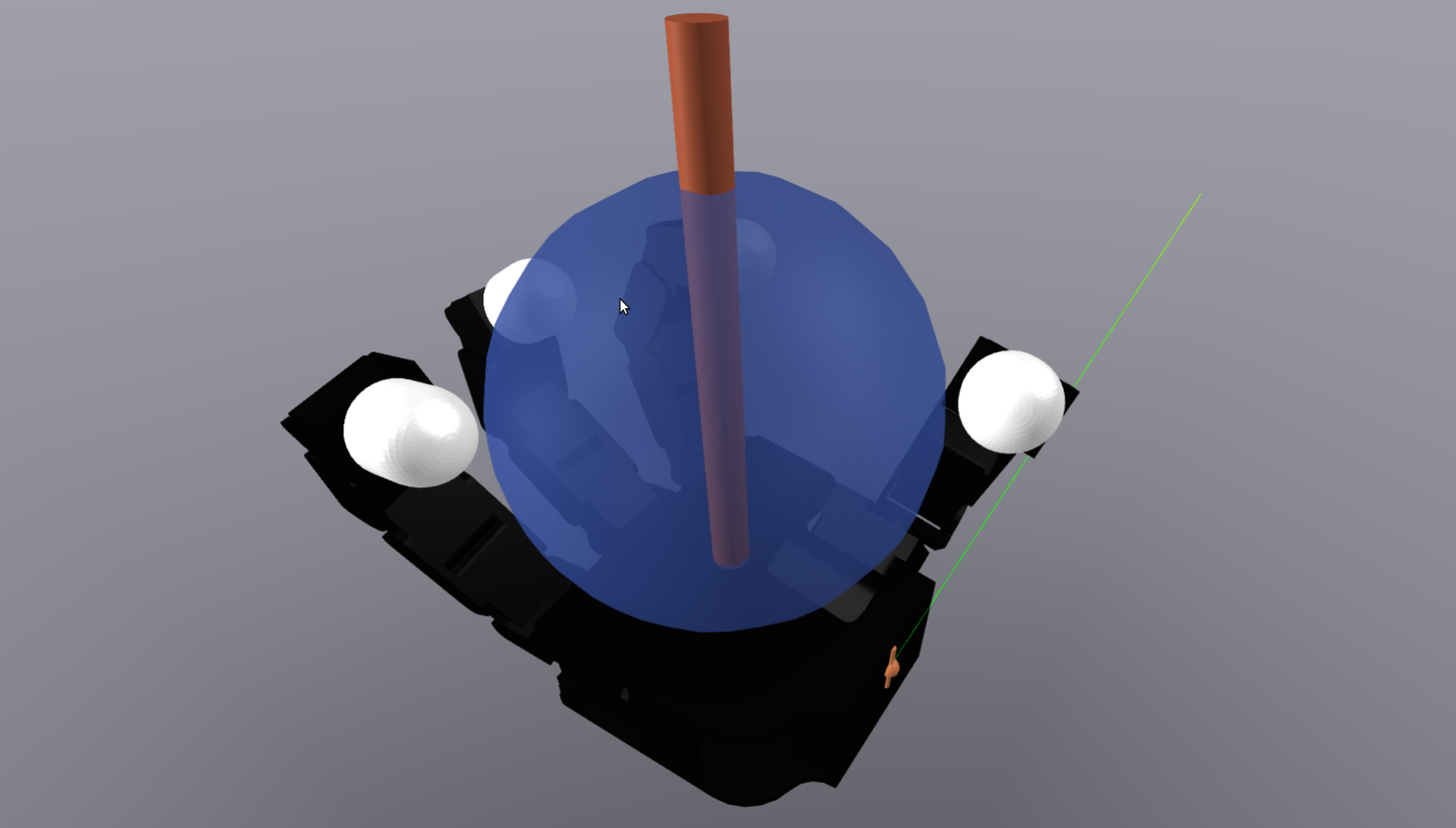}
            \vspace{-0.2cm}
            \label{fig: rotz_system_testbench}
        \end{minipage}
    }
    \subfigure[\textbf{Free}]{
        \begin{minipage}{4cm}
            \centering    
            \includegraphics[width=4cm]{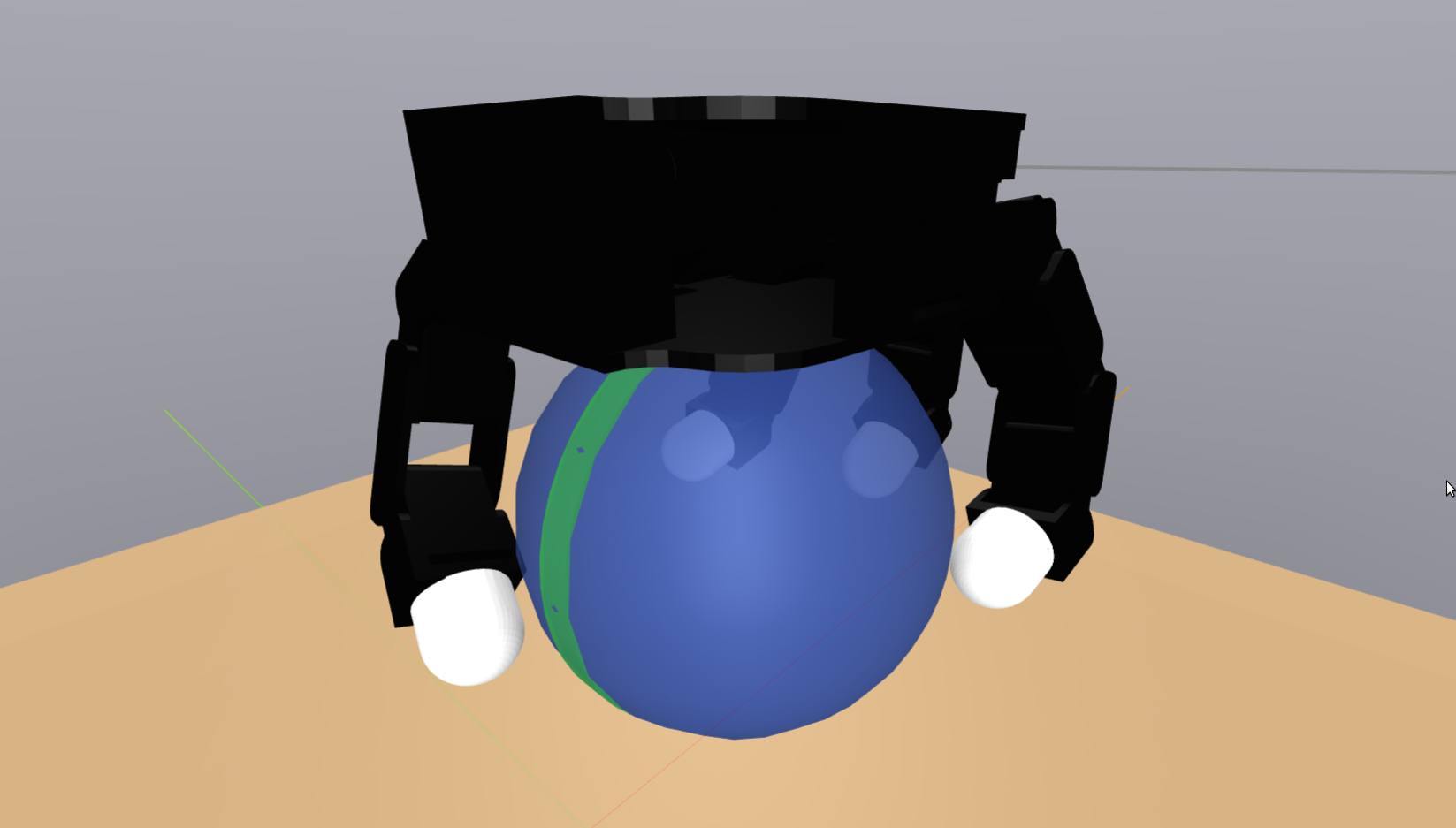}
            \vspace{-0.2cm}
            \label{fig: free_system_testbench}
        \end{minipage}
    }
    \vspace{-0.5cm}
    \caption{Two in-hand manipulation systems used in the experiments. The object being manipulated is painted in blue. (a) The orange cylinder represents a damped hinge. (b) The brown body represents the supporting plane.}
    \label{fig: test_cases}
    \vspace{-0.6cm}
\end{figure}

We carry out simulations and ablation studies to validate the proposed method. We use the CQDC model implementation from the quasi-static simulator \cite{Pang2021ACQ} proposed by Pang et al. The DDP algorithm is implemented using Python bindings of the Crocoddyl library \cite{mastalli20crocoddyl}. Simulations of the complete system are performed using Drake toolbox \cite{drake}, and the code base of \cite{Kurtz2023InverseDT} with hydroelastic contacts.
Two types of in-hand manipulation systems are used to evaluate the proposed method, as shown in Fig.~\ref{fig: test_cases}. We use a four-finger dexterous hand Allegro Hand with a fixed wrist and 16 actuated joints. The object to be manipulated could either rotate perpendicular to the palm (\textbf{RotZ}, $n_q=17$) or move freely while being supported by the table (\textbf{Free}, $n_q=23$)
\begin{figure}[!b]
    \vspace{-0.5cm}
    \centering
    \includegraphics[width=1\linewidth]{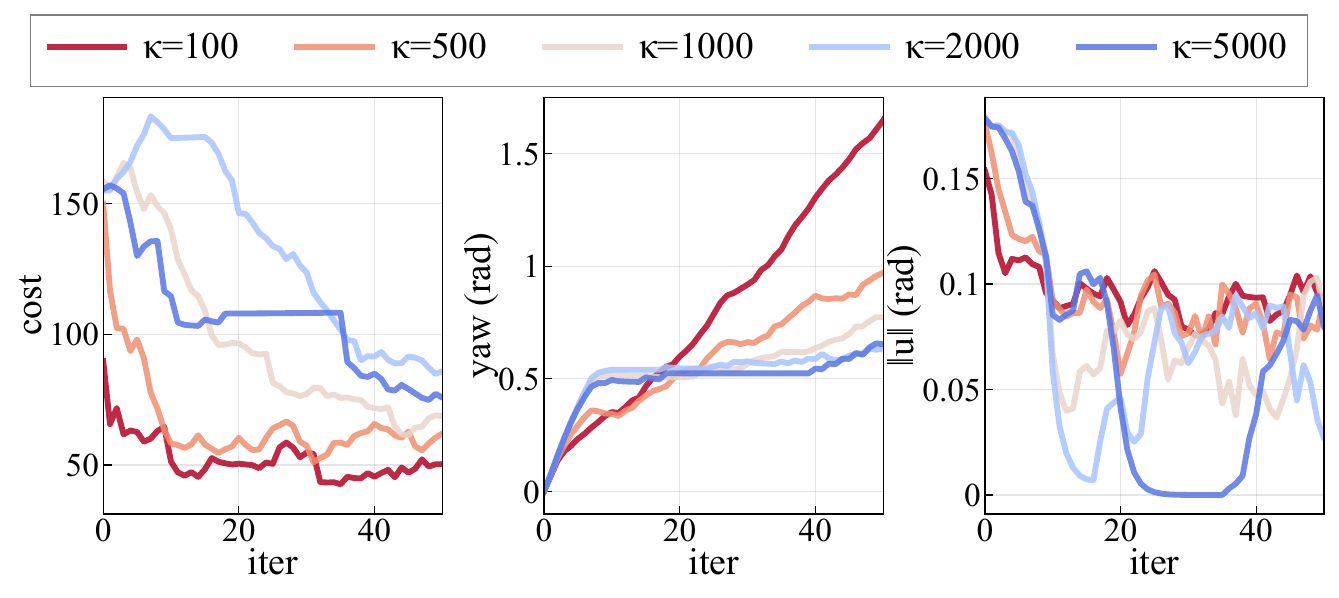}
    \vspace{-0.8cm}
    \caption{The relationship between $\kappa$ and optimization results. The cost of DDP, the object's yaw angle, and the control input norms are plotted for 50 steps. Warmer colors indicate a stronger smoothing effect.}
    \label{fig: kappa_change}
\end{figure}
%

\subsection{Long-Horizon Motion Generation with Rich Contacts} \label{sec: motion_generation_results}

\begin{figure}[!t]
    \centering
    \subfigure[Cylinder]{
        \begin{minipage}{1\linewidth}
            \centering
            \includegraphics[height=2.5cm]{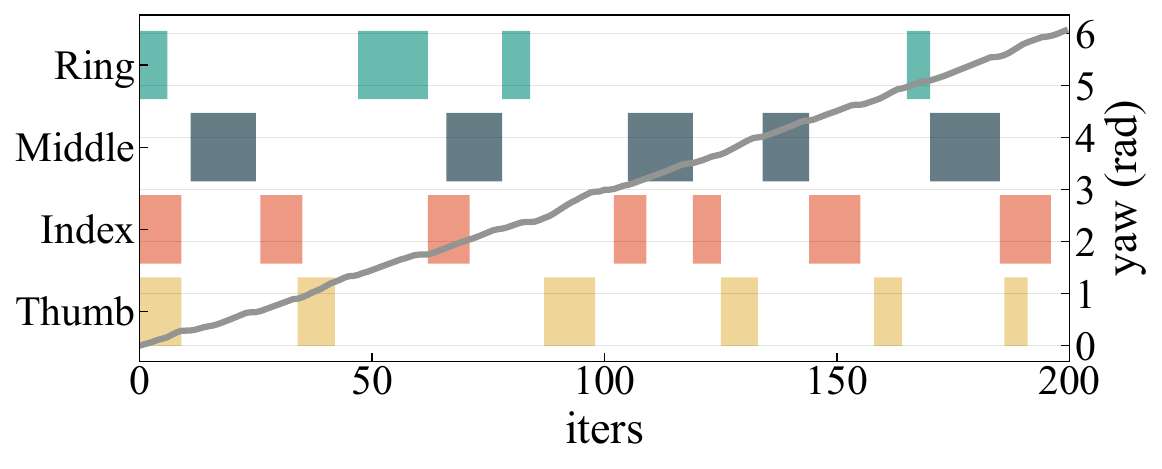}
            \hspace{-0.2cm}
            \includegraphics[height=2.5cm]{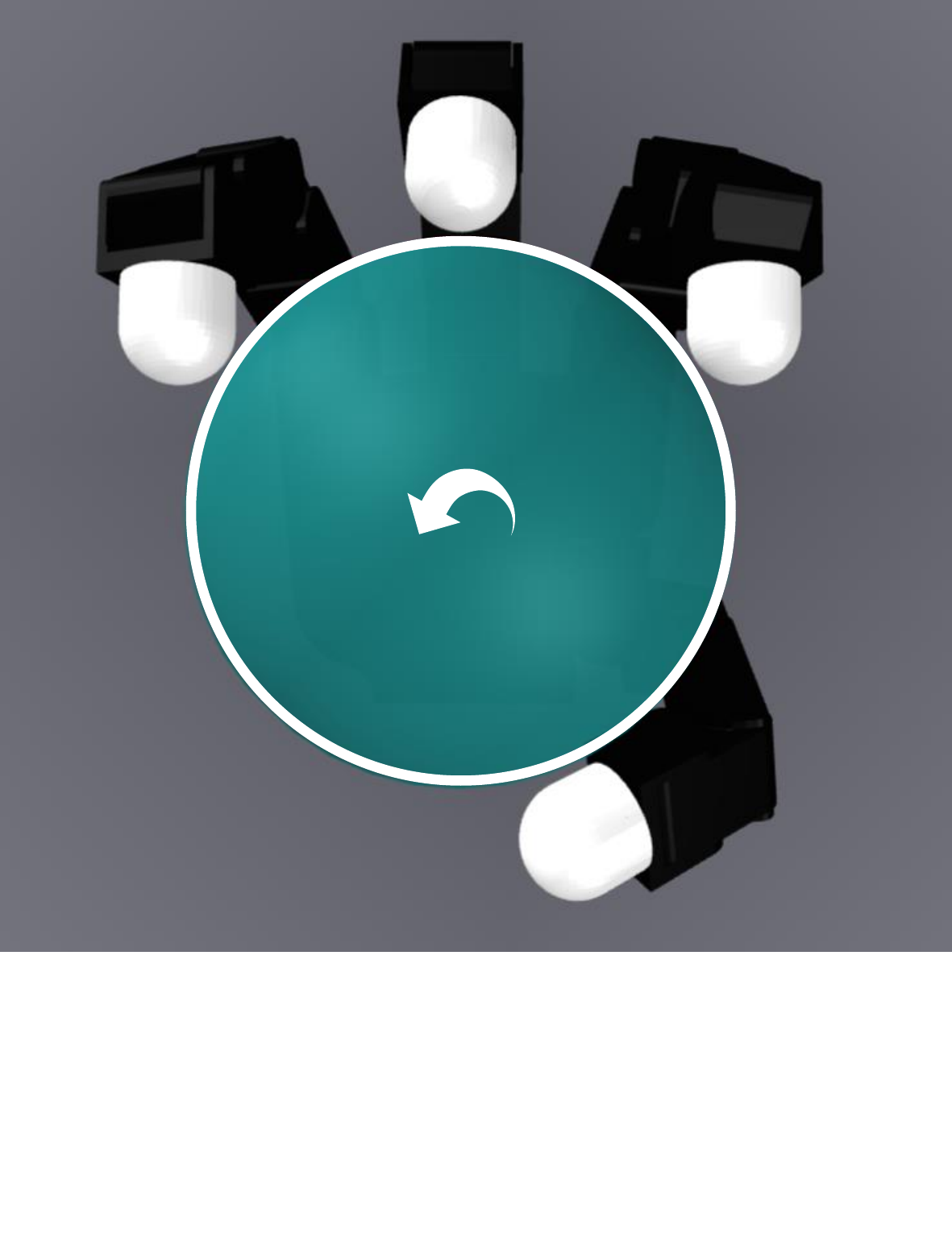}
        \end{minipage}
    }
    \subfigure[Cuboid]{
        \begin{minipage}{1\linewidth}
            \centering
            \vspace{-0.3cm}
            \includegraphics[height=2.5cm]{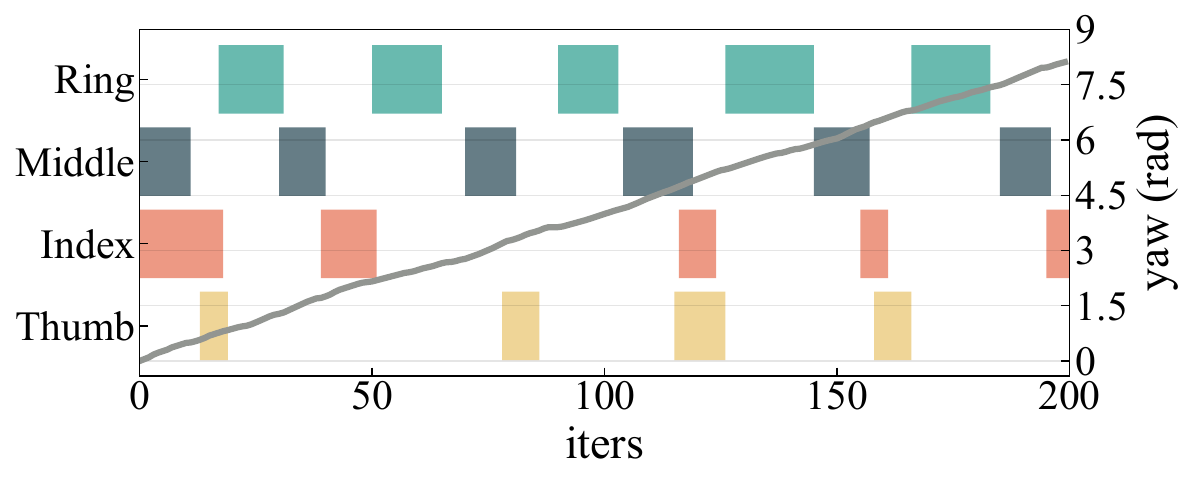}
            \hspace{-0.2cm}
            \includegraphics[height=2.5cm]{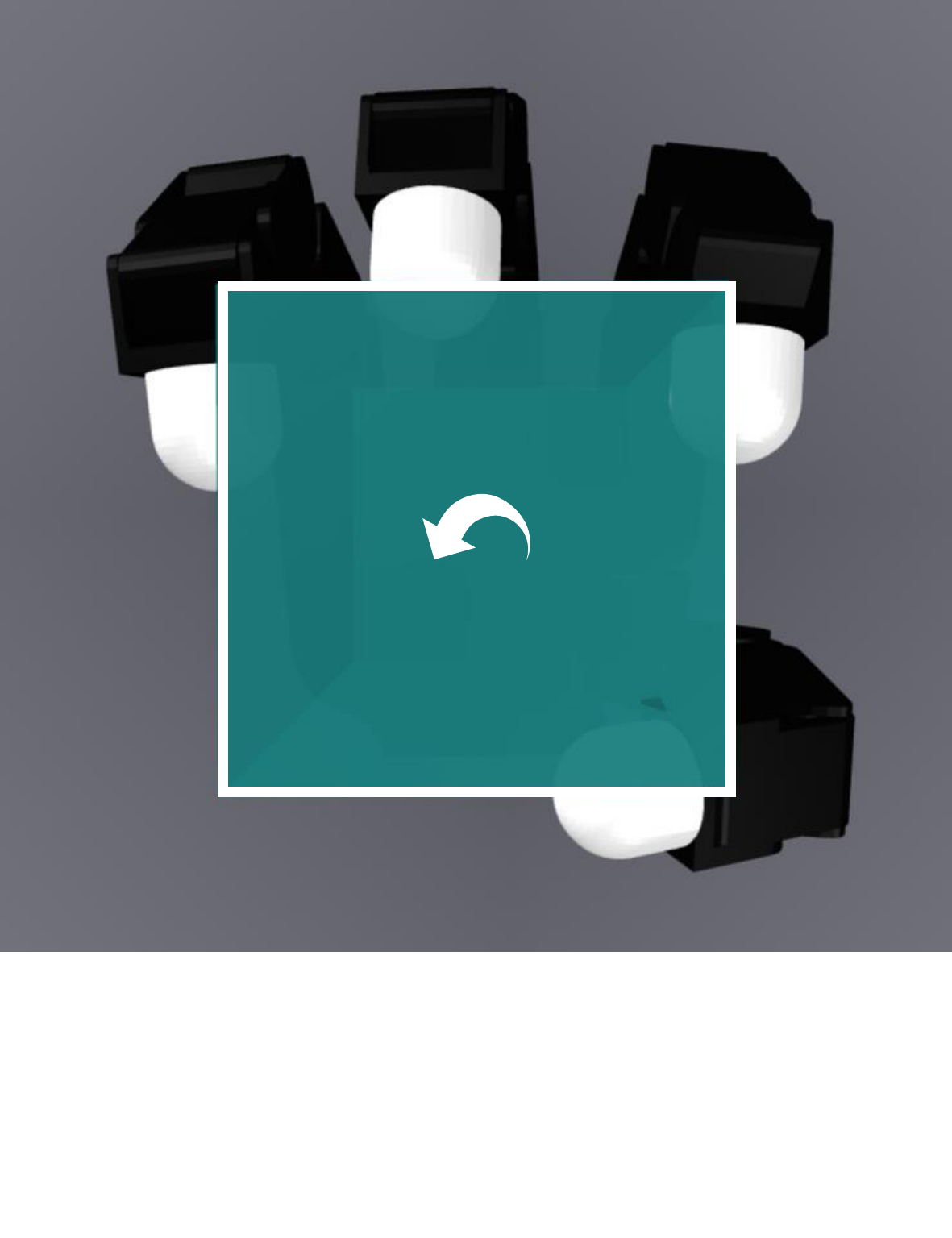}
        \end{minipage}
    }
    \vspace{-0.5cm}
    \caption{The finger-gaiting of the in-hand manipulation on two different objects, with a duration of 200 iterations. Shaded areas indicate contacts of each finger. The object yaw angle curves are plotted in grey.}
    \label{fig: finger_gaiting}
    \vspace{-0.6cm}
\end{figure}

%
We use $\kappa=100$ (the same notation as \cite{Pang2022GlobalPF}) to analytically smooth (\ref{eq: high_level_differentiable_dynamics}) throughout the paper, where a smaller value indicates a stronger smoothing effect. We set the input bounds as $\SI{0.05}{rad}$ for the \textbf{RotZ} system and $\SI{0.2}{rad}$ for the \textbf{Free} system, with the discretization step $h=\SI{0.1}{s}$ and the horizon length of $\SI{10}{}$.
First, we demonstrate the importance of using smoothed forward dynamics for contact-rich manipulation. We command the object to rotate at $\SI{0.79}{rad/s}$ and change $\kappa$ in the forward dynamics computation. As shown in Fig.~\ref{fig: kappa_change}, larger $\kappa$ leads to poorer convergence and less effective manipulation of the object. The yaw angles increase rapidly when $\kappa \geq \SI{1000}{}$ due to initial finger motions solved from warm-start trajectories. However, they quickly evolve into vibrating motions or even stop moving after 20 iterations without enough smoothing. Smoothed forward dynamics ($\kappa \leq 500$) enables the finger-gaiting generation and achieves constant rotation speed instead of getting stuck in a local optimum. Thus, we prove it impossible to remedy high-level modeling errors through a more realistic forward pass. Hence, a low-level controller is necessary.
Second, we test the generalization ability on different objects. As shown in Fig.~\ref{fig: teaser} and Fig.~\ref{fig: finger_gaiting}, a sphere with the radius of $\SI{6}{cm}$ and a cuboid with the side length of $\SI{7}{cm}$ are manipulated in the \textbf{RotZ} environment. The shaded areas indicate contact events ($\phi<\SI{3}{mm}$) of four fingers, where the finger-gaiting is auto-generated to accomplish the long-horizon manipulation. The sphere rotates at $\SI{0.29}{rad/s}$ and the cuboid rotates at $\SI{0.4}{rad/s}$. Our method seamlessly generalizes to different geometries without pre-training, which shows a potential advantage over learning-based approaches. See the attached video for details.
\begin{figure}[!b]
    \vspace{-0.7cm}
    \centering
    \subfigure[Rotation]{
        \hspace{-0.3cm}
        \begin{minipage}{4.0cm}
            \centering
            \includegraphics[height=2.95cm]{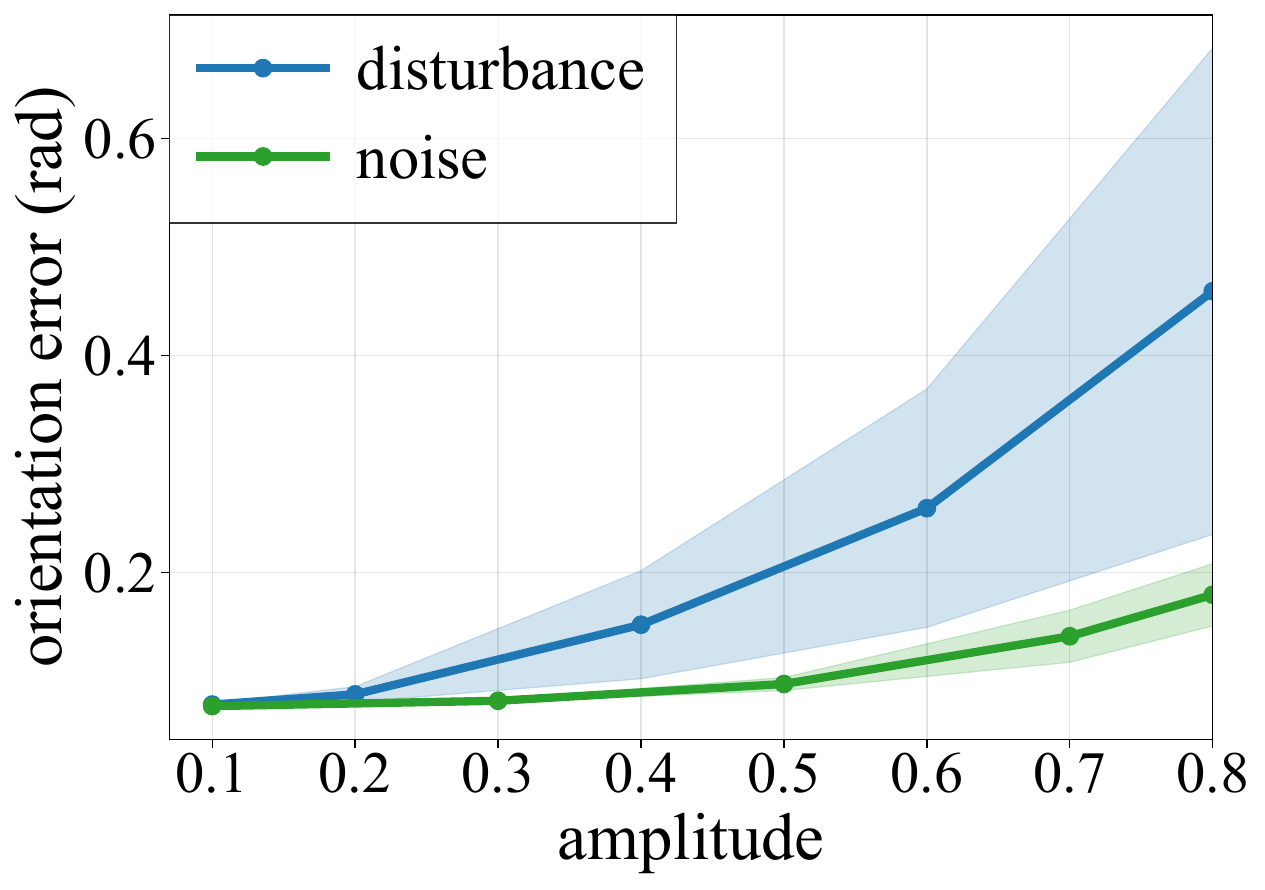}
            \vspace{-0.4cm}
        \end{minipage}
    }
    \subfigure[Position]{
        \hspace{-0.2cm}
        \begin{minipage}{4.0cm}
            \centering    
            \includegraphics[height=2.95cm]{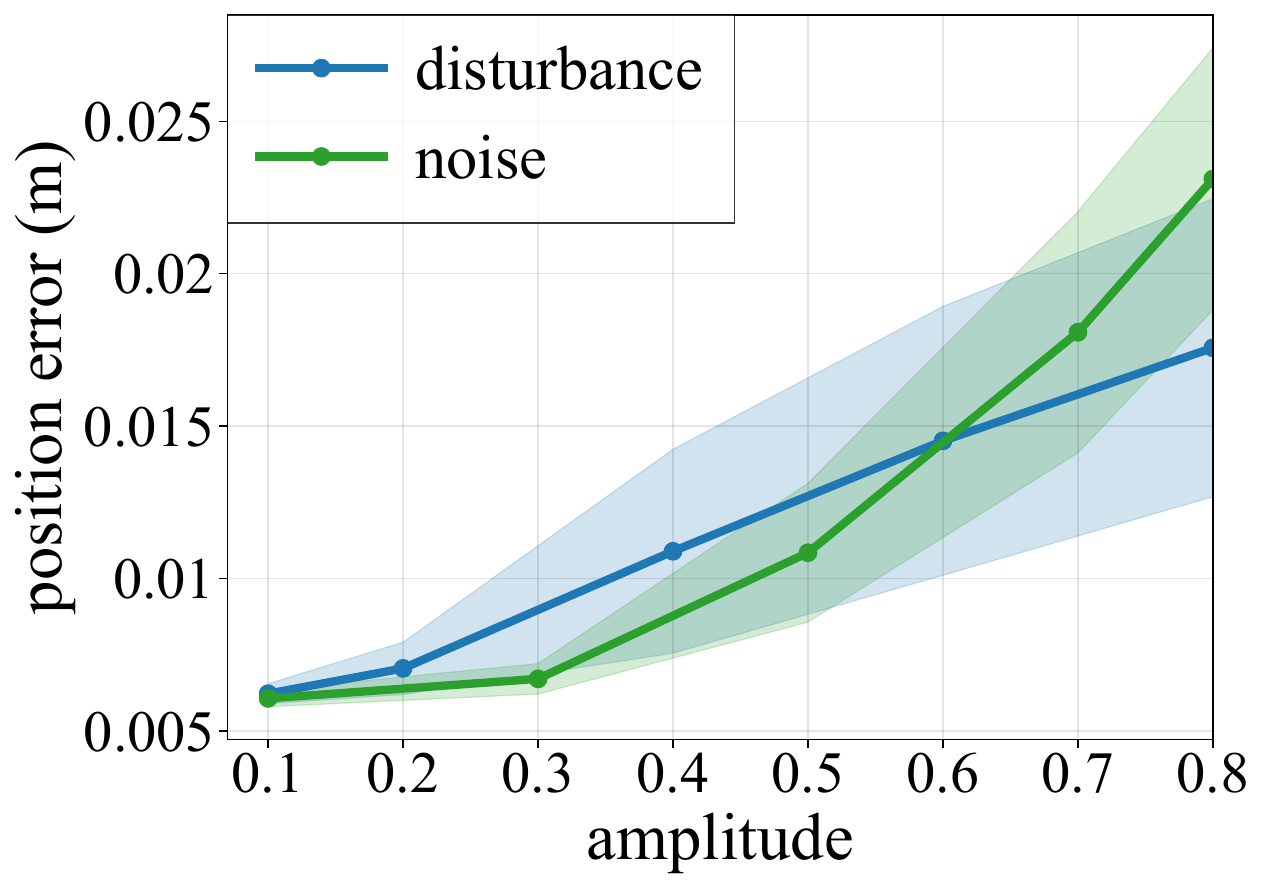}
            \vspace{-0.4cm}
        \end{minipage}
    }
    \vspace{-0.2cm}
    \caption{The rotation and position error curves with the increasing amplitude of noise/disturbance. The mean values (solid lines) and half standard deviation ranges (shaded areas) are shown.}
    \label{fig: robustness_results}
\end{figure}

%
\begin{figure*}[!t]
    \centering
    \subfigure[Valve]{
        \begin{minipage}{1\linewidth}
            \centering
            \includegraphics[height=2.8cm]{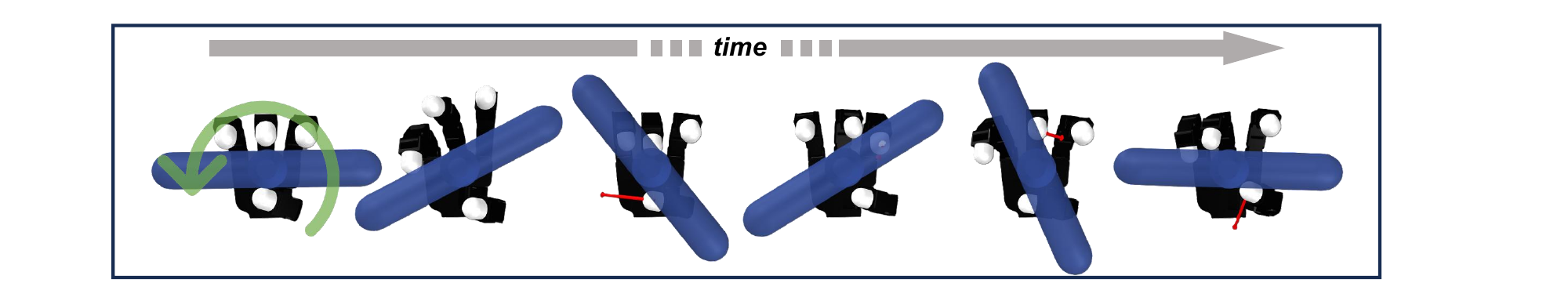}
            \hspace{-0.5cm}
            \includegraphics[height=2.8cm]{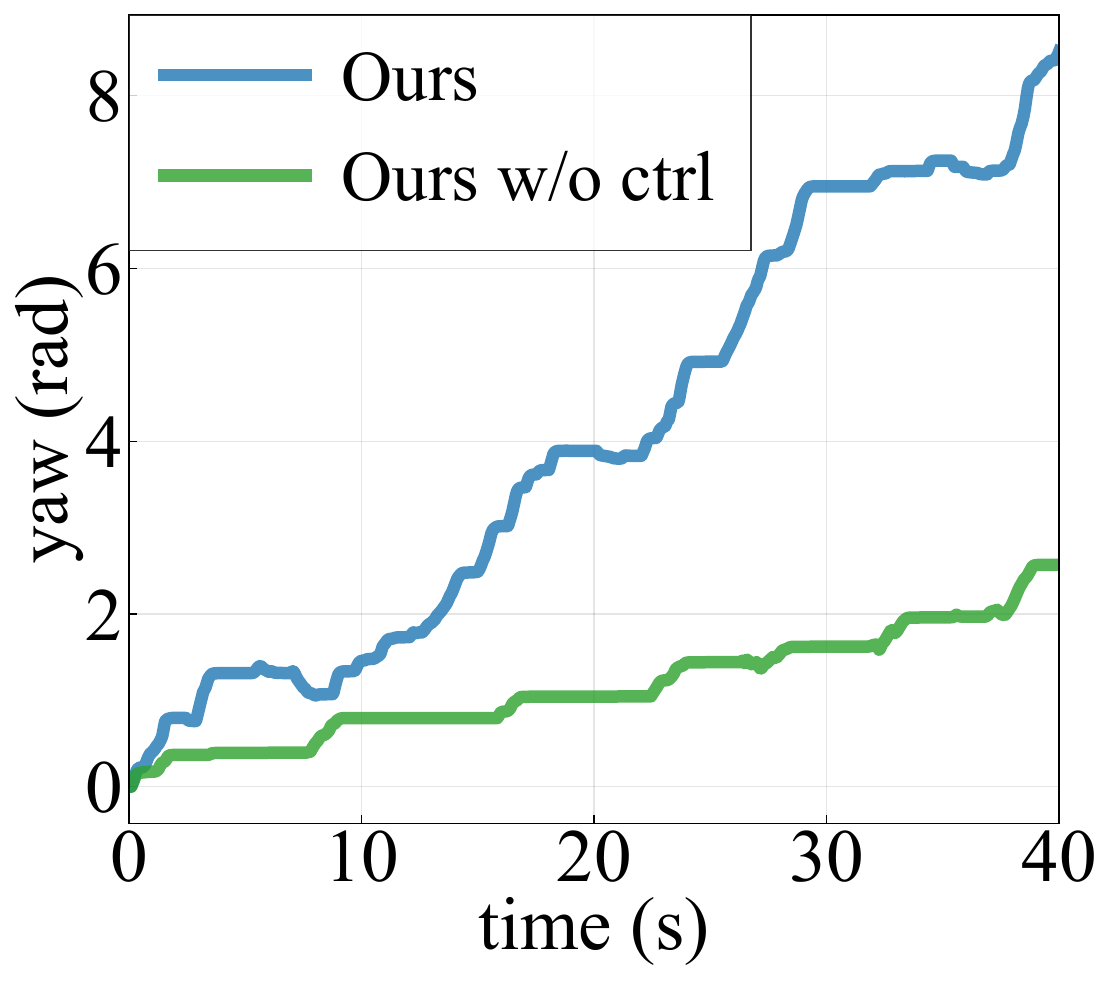}
            \vspace{-0.3cm}
            \label{fig: valve_rotation_snapshots}
        \end{minipage}
    }
    \subfigure[Sphere]{
        \begin{minipage}{1\linewidth}
            \centering
            \vspace{-0.32cm}
            \includegraphics[height=2.7cm]{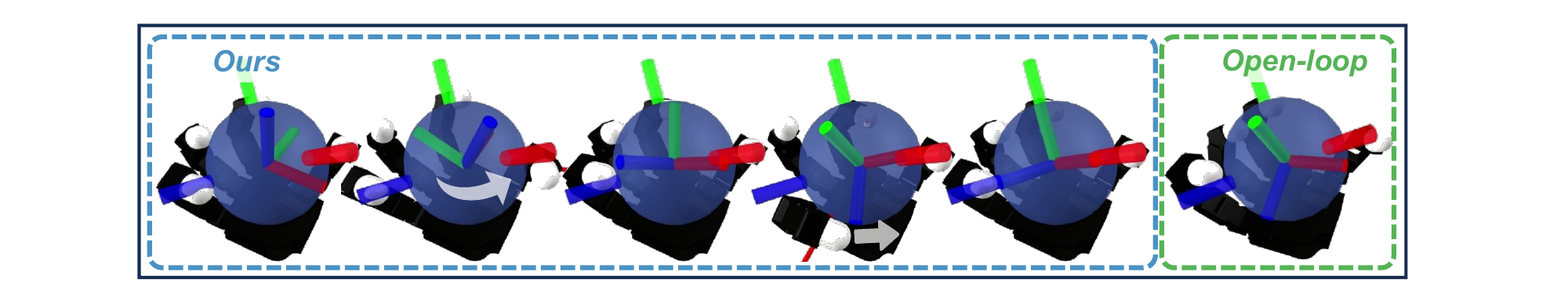}
            \includegraphics[height=2.8cm]{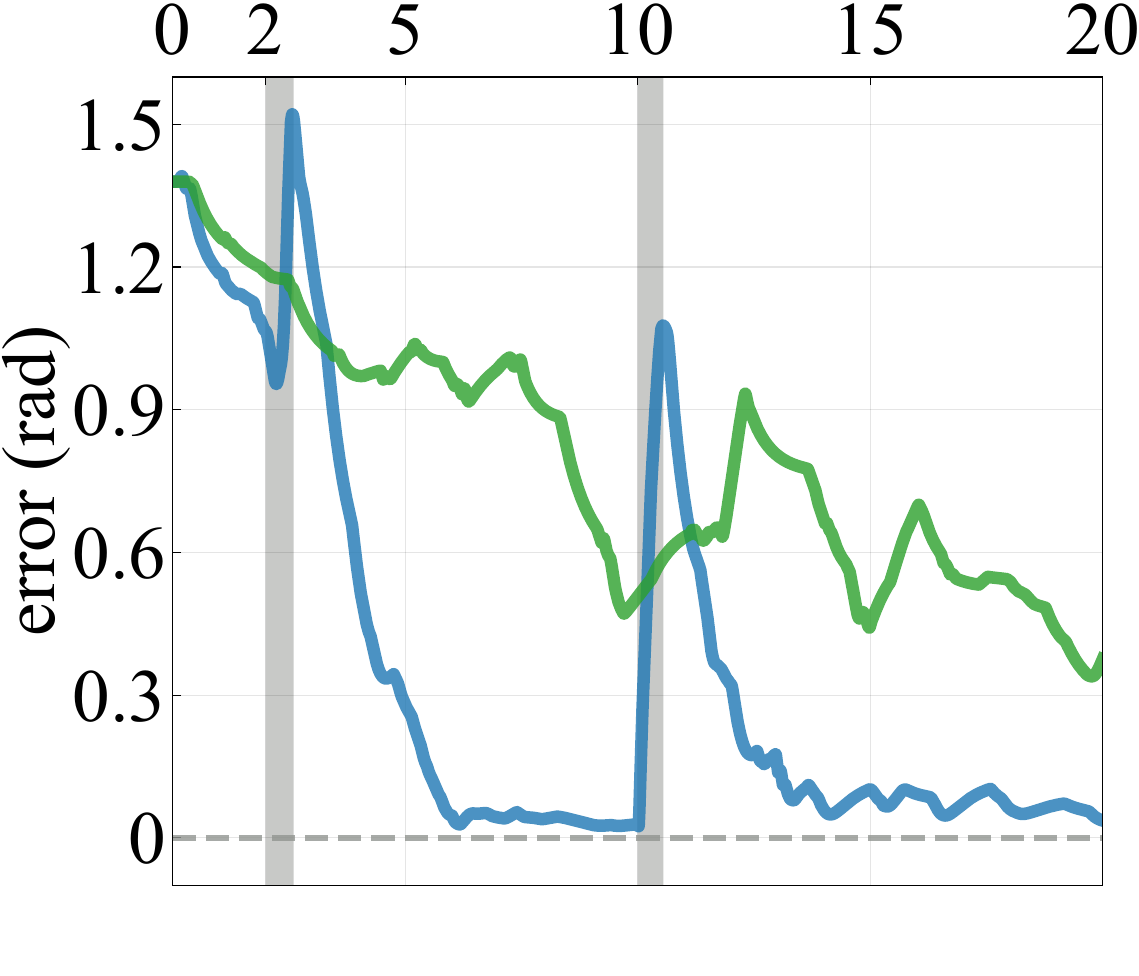}
            \vspace{0.1cm}
            \label{fig: sphere_rotation_snapshots}
        \end{minipage}
    }
    \vspace{-0.5cm}
    \caption{Snapshots of the long-horizon (a) valve rotation task and (b) sphere rotation task carried out in the Drake simulator. Red arrows represent the contact forces. (a) The blue capsule represents the valve and rotates around the vertical axis. The fingers perform side-pushes and step over the valve in succession. (b) External disturbances are applied on the object and ring finger, as the white arrows indicate. The final state of the open-loop baseline is also shown. Object yaw angle and error curves are shown on the right.}
    \label{fig: snapshots_and_error_curves}
    \vspace{-0.25cm}
\end{figure*}
Third, we further evaluate the robustness of our method under external disturbances or internal noises, which are assumed to obey Gaussian distributions, with the standard deviation $\sigma$. The sphere is commanded to rotate around the vertical axis in the \textbf{Free} system. To simulate the disturbances, we apply numerical perturbations once to the Euler angles and change $\sigma$ (i.e., the magnitude) from $\SI{0.1}{rad}$ to $\SI{0.8}{rad}$. To simulate the noises, at every time step, we perturb the dimensions of hand, object position, and quaternion in $\bm{x}$, with $\sigma=\SI{0.03}{rad},\SI{1}{mm},\SI{0.01}{}$, respectively. We define the amplitude as the multiples of $\sigma$. We run 100 random experiments for each parameter and record the average tracking error within 400 steps. As shown in Fig.~\ref{fig: robustness_results}, our method could recover from various disturbances and noises. Robustness is essential for long-horizon manipulation under sensor noises, control delays, and external disturbances. The orientation error explodes under large disturbances since the system needs a longer time for recovery.
%

\subsection{Robust Execution with Contact Controller}  \label{sec: contact_control_results}

%
\subsubsection{Simple tasks}  \label{sec: simple_tasks_low_level}
We choose three baselines for comparison. The first one executes generated trajectories in an open-loop fashion (Ours w/o ctrl), which resembles the motion planning in \cite{Pang2022GlobalPF} with online re-planning. The second one utilizes feed-forward torques $\bm{\tau}_{ff}=\bm{\mathcal{J}}^{\top}\bm{\mathcal{F}}_{e,ff}$ as (\ref{eq: joint_level_control_law}) (FF+torque), which is the quasi-dynamic variant of \cite{Kurtz2023InverseDT}. The third one maps feed-forward torques to desired positions, where $\bm{q}_d$ is replaced by $\bm{q}_d+\bm{\tau}_{ff}/\bm{K}_{\text{ctrl}}$ in (\ref{eq: joint_level_control_law}) (FF+pos), as suggested by \cite{Kim2023ContactImplicitMC}. 
As for our feedback design, low-cost tactile sensors \cite{Lambeta2020DIGITAN} can be used to obtain contact forces 
in hardware.
A slippage metric 
is adopted from \cite{Kim2023ContactImplicitMC} to indicate the occurrence of missed or ineffective contacts.
%

%
\begin{table}[!t]
    \centering
    \vspace{-0.3cm}
    \caption{Controller Performance in the RotZ System.}
    \vspace{-0.2cm}
    \label{tab: performace_low_level_ctrl}
        \begin{tabular}{@{}c|ccc@{}}
            \toprule
            Methods          & \begin{tabular}[c]{@{}c@{}c@{}}Average\\ slippage\\ ($\times 10^{-3}$)\end{tabular} & \begin{tabular}[c]{@{}c@{}c@{}}Average\\ rotation speed\\ (rad/s)\end{tabular} & \begin{tabular}[c]{@{}c@{}c@{}}Average\\ joint velocity\\ (rad/s)\end{tabular} \\ \midrule
            Ours              & \textbf{3.26}   & 0.30  & 0.93     \\
            Ours w/o ctrl     & 8.04   & 0.05  & 0.83     \\
            FF+torque         & 9.47   & 0.85  & 1.17     \\
            FF+pos            & 8.80   & 0.85  & 1.18      \\ \bottomrule
        \end{tabular}
    \vspace{-0.6cm}
\end{table}
We simulate the \textbf{RotZ} system in Drake for $\SI{10.0}{s}$ with the four controllers. The reference generator outputs references at $\SI{20}{Hz}$, and the contact controller runs at $\SI{100}{Hz}$. We record the average slippage when $\phi<\SI{1}{mm}$, the object's average rotation speed, and the hand's average speed in joint space.
As depicted in Table \ref{tab: performace_low_level_ctrl}, our controller achieves the lowest average slippage by maintaining enough desired force. As a result, the average rotation speed increases significantly compared with open-loop execution, which indicates that more effective contacts are established with our method. The two feed-forward controllers perform similarly and rotate the sphere faster with higher slippage. This is because the feed-forward torques are converted into displacements due to joint impedance. Thus, feed-forward controllers often result in larger deviations from planned motions and faster joint velocities once the contact is lost, as seen in the last column of Table \ref{tab: performace_low_level_ctrl}. In contrast, our controller balances tracking planned motions and making contacts, thus generating more stable motions.
%

%
%

%
\subsubsection{Complex tasks}  \label{sec: complex_tasks_low_level}
We further evaluate the contact controller and the proposed method in more complex cases. These experiments also show potential applications in the real world. 
The first case is rotating the valve, where $n_q=17$. In this case, the fingers should switch between pushing aside and stepping over the valve to accomplish the long-horizon manipulation, shown in Fig.~\ref{fig: valve_rotation_snapshots}. To our knowledge, such long-horizon manipulation is almost only studied in the machine-learning literature. We find it helpful to interpolate the nominal finger configuration $\bm{q}_{o,\text{ref}}$ between two grasping poses of different fingertip heights. Otherwise, a longer horizon length is needed to avoid local optimums, which degrades control frequency. As shown in Fig.~\ref{fig: valve_rotation_snapshots}, the fingers sometimes need to exert forces below the valve, which is sensitive to contact loss. Our proposed contact controller effectively avoids such problems, as shown in the curves and the attached video.
The second case is rotating the sphere in place, where $n_q=19$. The sphere should be aligned with the target orientation in SO(3). With our proposed controller, we exert spatial force disturbances on the object during $\SI{2.0}{s}\sim\SI{2.5}{s}$ and on the ring finger during $\SI{10.0}{s}\sim\SI{10.5}{s}$.
As shown in Fig.~\ref{fig: sphere_rotation_snapshots}, the long-horizon manipulation fails without the contact controller due to contact loss and unintentional contact.
Hence, the importance of our contact controller is proved.
We also simulate the \textbf{Free} system in Fig.~\ref{fig: free_system_testbench} and successfully rotate the sphere for more than $270^\circ$. However, the rolling contacts result in highly dynamic movements of the object, which violates the quasi-dynamic assumption and degrades the performance. We temporarily increase the damping of the object. As suggested by \cite{Pang2022GlobalPF}, designing special high-level objectives and enforcing the system to be quasi-dynamic at a low level could help to solve this problem.
We leave them for future work.

\vspace{-5pt}
\section{Conclusions}
This paper discusses the task of long-horizon in-hand manipulation. First, the reference generator computes joint trajectories and reaction forces based solely on desired motions. Then, the contact controller, which includes tactile feedback, tracks joint movements and maintains planned contacts even in the presence of modeling errors. The experiments demonstrate that the proposed method achieves similar dexterity to learning-based approaches and generalizes to different objects without pre-training. With the reference generator, our method achieves long-horizon manipulation without a separate planning procedure by recalculating trajectories in real-time. In addition, the contact controller prevents contact loss and enhances robustness to disturbances. Future work will focus on hardware implementation with tactile sensors and improving manipulation performance through hand-arm coordination.
%






{
\small
\bibliographystyle{IEEEtran}
\bibliography{ref}
}


\end{document}